\documentclass{article}
\usepackage{graphicx}
\usepackage{CJKutf8}
\usepackage{tabularx}
\usepackage{amssymb}
\usepackage[T1]{fontenc}
\usepackage{neurips24}
\usepackage{url}
\usepackage[dvipsnames]{xcolor}
\usepackage{amsmath}
\usepackage{multirow}
\usepackage{makecell}
\usepackage{vcell}
\newcommand{\paratitle}[1]{\vspace{0.8ex}\noindent \textbf{#1}}

\title{Open-domain Implicit Format Control for Large Language Model Generation}

\author{
  Yiqun Yao\textsuperscript{1\textdagger},
  Wenjia Ma\textsuperscript{2\textdagger},
  Xuezhi Fang\textsuperscript{1}, 
  Xin Jiang\textsuperscript{1},
  Xiang Li\textsuperscript{1}, 
  Xuying Meng\textsuperscript{3},\\
  \textbf{Peng Han\textsuperscript{4}, Jing Li\textsuperscript{5}, Aixin Sun\textsuperscript{6}, Yequan Wang\textsuperscript{1$*$}}\\
  $^{1}$Beijing Academy of Artificial Intelligence, Beijing, China\\
  $^{2}$AstralForge AI Lab, Beijing, China\\
  $^{3}$Institute of Computing Technology, Chinese Academy of Sciences, Beijing, China\\
  $^{4}$University of Electronic Science and Technology of China, Chengdu, China\\
  $^{5}$Harbin Institute of Technology, Shenzhen, China\\
  $^{6}$School of Computer Science and Engineering, Nanyang Technological University, Singapore
}

\begin{document}

\maketitle

\begin{abstract}
Controlling the format of outputs generated by large language models (LLMs) is a critical functionality in various applications. Current methods typically employ constrained decoding with rule-based automata or fine-tuning with manually crafted format instructions, both of which struggle with open-domain format requirements. To address this limitation, we introduce a novel framework for controlled generation in LLMs, leveraging user-provided, one-shot QA pairs. This study investigates LLMs' capabilities to follow open-domain, one-shot constraints and replicate the format of the example answers. We observe that this is a non-trivial problem for current LLMs. We also develop a dataset collection methodology for supervised fine-tuning that enhances the open-domain format control of LLMs without degrading output quality, as well as a benchmark on which we evaluate both the helpfulness and format correctness of LLM outputs. The resulting datasets, named OIFC-SFT, along with the related code, will be made publicly available at \url{https://github.com/cofe-ai/OIFC}.

\end{abstract}

\section{Introduction}
Large Language Models (LLMs) are extensively utilized across academic, industrial, and everyday settings \cite{GPT-4,gemini,claude,grok,sparks}. Rigorous user surveys \cite{we} indicate a strong demand for well-structured outputs from LLM-generated content. However, even if meticulously prompted, current models frequently yield undesirable responses, necessitating labor-intensive post-processing during mass API usage. Providing numerous examples \cite{gpt3, cot} partially addresses this issue, yet it leads to inefficient token usage due to lengthy prefixes. Furthermore, accessing few-shot demonstrations for open-domain implicit formats proves challenging, which limits the scaling of related training data. Consequently, there remains a significant disparity between the need for open-domain format control and the current alignment capabilities of LLMs.

Specifically, while the alignment data often include some widely-used formats (e.g. JSON, YAML, markdown lists, bullet points, etc.), the model only learns from these predefined formats and may not meet unique user requirements. On the other hand, if the users explicitly specify the required formats in prompt, the model occasionally overlooks them, or adheres to them while inserting unexpected explanations around the output. In summary, current models need further tuning to adequately meet the demands for user-driven, open-domain format control.

A pivotal question is how users should articulate their needs. Research shows that preferences vary between web UI and textual descriptions, depending on the task \cite{we}. However, existing frameworks rely on predefined, explicit format sets, which limits their flexibility. In contrast, our proposed framework utilizes \textit{implicit} format descriptions derived from one-shot examples provided by users. This approach addresses the challenges of specifying highly complex requirements and bridges the gap between user demands and predefined formats.  Notably, even plain text lacking a clear format can be utilized for format control, with the expectation that the model will adopt a similar \textit{implicit} tone in its responses.

In our evaluation and case studies, we found that the user-provided examples are often disregarded, underscoring the complexity of empowering models with such abilities. We developed a data collection methodology, resulting in a training dataset and a testing benchmark tailored to our proposed framework. Through supervised fine-tuning (SFT), we observed notable improvements in open-domain format control with negligible fluctuations in the helpfulness of model responses. Another advantage of our one-shot framework is that even SFT data with human-annotated responses deemed ``low quality'' can enhance format control abilities with minimal impact on helpfulness, as even poorly-formed responses adequately demonstrate certain implicit formats. This feature significantly eases the scalability of our data collection method.

To summarize, our contributions include: (i) we design a new framework for Open-domain Implicit Format Control (OIFC); (ii) we develop a dataset collection pipeline that facilitates the training and evaluation for our framework; the resulting OIFC-SFT dataset and related code will be open-sourced; (iii) our experimental results support the effectiveness of these methodologies.

\section{One-shot Implicit Format Control}
\subsection{Definition}
We define One-shot Implicit Format Control (OIFC) as an LLM decoding framework formulated as:
\begin{equation}
    y = f(x; \{p, q_{one\_shot}, r_{one\_shot}\}),
    \label{eq1}
\end{equation}
where the model response $y$ to the query $x$ is expected to adhere to the implicit format requirements exemplified by the one-shot example response $r_{one\_shot}$. Here, $p$ represents an optional prompt that includes textual descriptions and the roles of each component; $q_{one\_shot}$ is an optional one-shot example query provided by the user or selected from a set of candidate queries that are similar by nature.

An example prompt in the OIFC format is demonstrated in Table \ref{tab:prompt_example}. In this case, the users are not required to describe the specific constrains on the chain-of-thought behavior (i.e., writing ``you should skip the step of computing (train length + tunnel length), and directly start from computing (total length / speed)''). Instead, they can simply provide a one-shot example that implicitly demonstrates the desired approach.
\begin{table*}[h]
  \centering
  \caption{Example of the OIFC-formatted prompt in our dataset.}
\footnotesize
\begin{CJK}{UTF8}{gbsn}{
\scalebox{0.83}{
    \begin{tabularx}{\textwidth}{c|X}
    \hline
    Chinese & \multicolumn{1}{m{10.5cm}}{以下是一对示例问题和示例回答。\textbackslash n【示例问题】\textbackslash n一列火车长200米，它以每秒10米的速度穿过200米长的隧道，从车头进入隧道到车尾离开隧道共需要多少秒?\textbackslash n【示例回答】\textbackslash n400÷10=40秒\textbackslash n共需要40秒。\textbackslash n\textbackslash n请回答以下问题，同时尽量保证回答的格式与上述【示例回答】相同。\textbackslash n【问题】\textbackslash n一辆汽车长150米，它以每秒20米的速度穿过100米长的桥梁，从车头进入桥梁到车尾离开桥梁共需要多少秒?}\\\hline
    English Translation & \multicolumn{1}{m{10.5cm}}{The following is a pair of example question and answer.\textbackslash n**Example Question**\textbackslash n A train is 200 meters long and moving at a speed of 10 meters per second. The train enters a 200-meter-long tunnel from the front, and it takes some time for the entire train to exit the tunnel from the back. How many seconds does it take for the train to completely pass through the tunnel? \textbackslash n**Example Answer**\textbackslash n 400 / 10 = 40 seconds. \textbackslash n The answer is 40 seconds. \textbackslash n \textbackslash n Please answer the following question with the closest format to the example answer. \textbackslash n**Question**\textbackslash n A car is 150 meters long and is traveling at a speed of 20 meters per second across a bridge that is 100 meters long. How many seconds does it take for the car to pass completely through the bridge from the front entering to the rear exiting?}\\
    \hline
    \end{tabularx}%

    }
  \label{tab:prompt_example}%
}\end{CJK}
\end{table*}%

\subsection{OIFC's Advantages in Open Domain}
It is straightforward to construct few-shot/one-shot SFT samples from existing NLP task datasets with closed-set answers, e.g. text classification, semantic parsing, etc. \cite{t5, flan, in-context-survey}. However, this approach encounters two significant challenges: (i) the diversity of formats does not increase with the amount of data, but only with the variety of tasks or manually crafted formats; (ii) the resulting data predominantly comprises explicitly converted formats, resulting in a scarcity of training samples for implicit formats. 

In contrast, within the framework of One-shot Implicit Format Control (OIFC) for open-domain applications, users can provide any text as a one-shot example of an implicit format, surpassing the limitations of predefined settings. Therefore, we aim to construct an OIFC Supervised Fine-Tuning (SFT) dataset using open-domain dialogues as sources. Here, the examples are derived from existing free-form responses, ensuring that format diversity naturally increases with the scale of the data. It is intuitive that replacing OIFC with few-shot alternatives would be counterproductive, as multiple free-form responses could conflict in terms of implicit formatting. This highlights the necessity for using \textit{one-shot} examples.

\subsection{Data Collection}
\label{sec:data-collection}

\begin{figure}[t]
    \centering
    \includegraphics[width=0.89\textwidth]{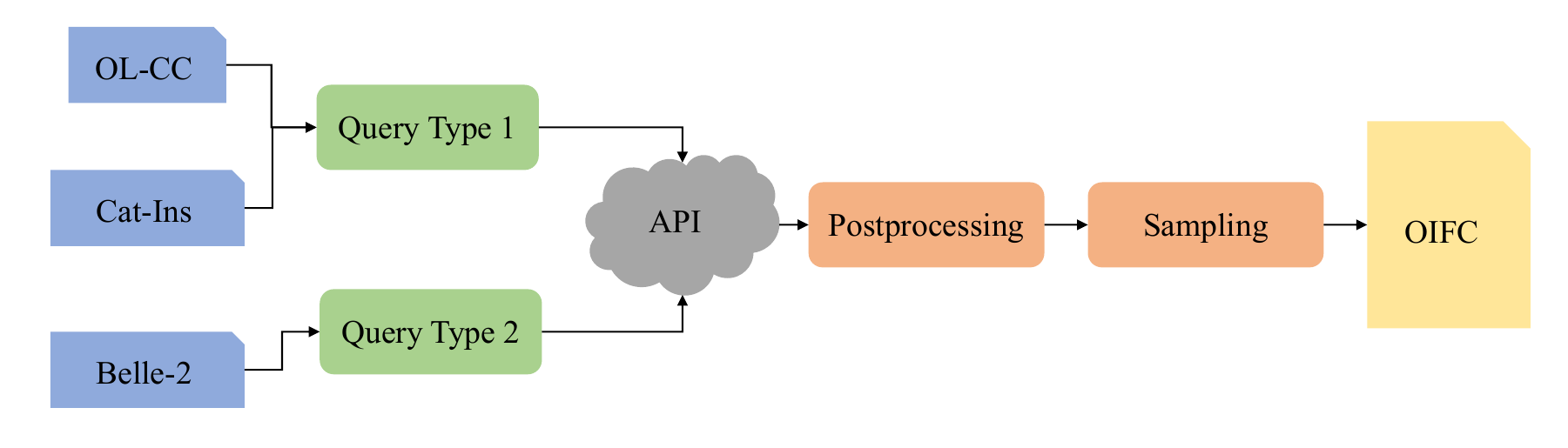}    
    \caption{Data collection pipeline of OIFC-SFT.}
    \label{fig:data_pipeline}
\end{figure}

\paratitle{Data Source.}
We aim to construct the OIFC-SFT dataset based on existing instruct tuning data. However, a vast majority of current open-sourced SFT data are sourced from proprietary LLM responses, e.g. GPT-4 \cite{GPT-4}. This reflects the default formats encoded in the proprietary LLMs, potentially limiting the response variety. To this end, we seek SFT data which involves intense human annotation and responses from earlier ``weaker'' models. The data construction pipeline is demonstrated in Figure \ref{fig:data_pipeline}. We focus on applications in Chinese. The source of OIFC-SFT data include:
\begin{itemize}
    \item \paratitle{OpenLabel - Chinese Conversations Dataset (OL-CC)}. OL-CC\footnote{\url{https://data.baai.ac.cn/details/OL-CC}} is a human-sourced instruction dataset featuring open-domain single-turn conversations. Human responses in OL-CC tend to be significantly shorter and sometimes less informative compared to those generated by GPT-like models. These features are beneficial as we are training a model to adhere to a user-provided, one-shot specific answer format, which is disentangled from the ability of ``generating helpful responses''.
    \item \paratitle{Categorized Instructions (Cat-Ins)}. To ensure the domain coverage of instructions, we complement with a dataset featuring similar response collection pipelines to OL-CC, whose instructions are further categorized into more than 100 classes by human annotators. We keep 10k samples across 96 categories, supplementing the OL-CC samples.
    \item \paratitle{Belle-2}. Belle-2\footnote{\url{https://huggingface.co/datasets/BELLE-2/train_3.5M_CN_With_Category}} is a typical open-sourced SFT dataset inspired by instruct enhancement methods such as Alpaca \cite{alpaca}. Given that responses in Belle-2 are sourced exclusively from ChatGPT, we primarily utilize their instructions rather than the responses.
\end{itemize}

\paratitle{Query Types.}
In order to construct samples in OIFC format, we incorporate three components: the intended instruct, a one-shot instruct and response pair, and the ground-truth response to the intended instruct as supervision. Depending on the characteristics of data sources, we design two types of queries to utilize the external API of advanced proprietary LLMs to obtain the necessary OIFC components.

\textit{Query Type 1.} Given an instruct-response pair $(\textcolor{red}{i_1}, \textcolor{red}{r_1})$ from the data source, we prompt the external model to first generate a similar instruct $\textcolor{blue}{i_2}$, and then answer $\textcolor{blue}{i_2}$ normally as if there is no prior context or format constraint, yielding a detailed response $\textcolor{green}{r_d}$. Finally, the model should restructure $\textcolor{green}{r_d}$ to adhere to the format of $\textcolor{red}{r_1}$ as much as possible, resulting in the ground-truth answer $\textcolor{blue}{r_2}$. After post-processing the API results, we fill Eq.\ref{eq1} with:

\begin{equation}
    \textcolor{blue}{r_2} = f(\textcolor{blue}{i_2}; \{p,\textcolor{red}{i_1}, \textcolor{red}{r_1}\}).
    \label{eq2}
\end{equation}
~~~~Query Type 1 is applied to process the OL-CC and Cat-Ins data sources. Typically, the generated instruct $\textcolor{blue}{i_2}$ is a reasonable question in the same domain as $\textcolor{red}{i_1}$, preserving the domain distribution in the original data sources. $\textcolor{green}{r_d}$ is designed to achieve high-quality responses leveraging the powerful chain of thoughts in modern LLM products. This ensures that the the ground-truth answers are more reliable even if the one-shot implicit format requires very short responses. By using the resulted $(\textcolor{blue}{i_2},\textcolor{blue}{r_2})$ as the main instruct-response pair and $(\textcolor{red}{i_1}, \textcolor{red}{r_1})$ as the one-shot example of implicit formats, this strategy both leverages the diverse response formats and alleviates the problems regarding response quality.

\textit{Query Type 2.} Beginning with an instruct $\textcolor{red}{i_1}$ from the data source, we directly prompt the external model to conceive two formats that differ maximally in both linguistic structures and sequence lengths. The model then directly answers $\textcolor{red}{i_1}$ in these distinct formats, yielding $\textcolor{blue}{r_1^1}$ and $\textcolor{blue}{r_1^2}$. Next, ask the model to generate a similar instruct $\textcolor{blue}{i_2}$, and answer it following the same formats, yielding $\textcolor{blue}{r_2^1}$ and $\textcolor{blue}{r_2^2}$. Finally, we fill them in Eq.\ref{eq1} and construct two OIFC-SFT samples:
\begin{align}
    \textcolor{blue}{r_2^1} &= f(\textcolor{blue}{i_2}; \{p,\textcolor{red}{i_1}, \textcolor{blue}{r_1^1}\}),\\
    \textcolor{blue}{r_2^2} &= f(\textcolor{blue}{i_2}; \{p,\textcolor{red}{i_1}, \textcolor{blue}{r_1^2}\}).
    \label{eq3}
\end{align}
~~~~In contrast to Query Type 1, Query Type 2 elicits the inherent variance of implicit formats in external model, using only the instruct $\textcolor{red}{i_1}$ as a seed. This inherent format distribution is in complement to the more natural format distribution as employed by Type 1, enhancing the model’s training by exposing it to a broader range of format possibilities. Additionally, this method facilitates the generation of multiple samples from the same instruction, which informs the model of the target ability in training. We use Query Type 2 with the Belle-2 instructions.

\begin{figure}[t]
    \centering
    \includegraphics[width=0.6\textwidth]{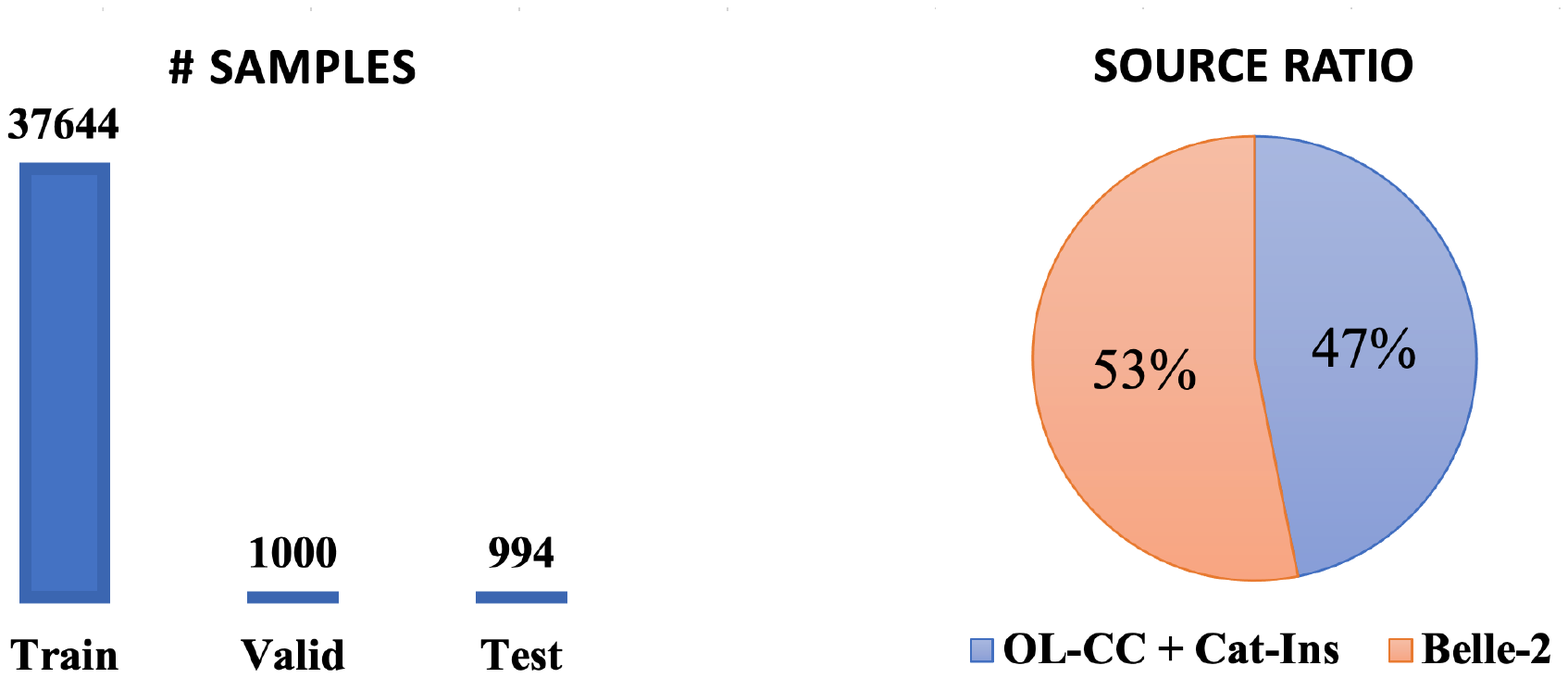}    
    \caption{Statistics for OIFC-SFT.}
    \label{fig:stats}
\end{figure}

\paratitle{Statistics.}
The split statistics and source mixing ratios of the resulting OIFC-SFT dataset is demonstrated in Figure \ref{fig:stats}.

\subsection{The OIFC Evaluation Protocol}
\paratitle{Benchmark Data.}
We utilize the OIFC test set, which comprises 994 samples, to form an in-distribution (ID) benchmark. Additionally, we employ Query Type 1 to process 500 samples from single-turn ShareGPT\footnote{\url{https://sharegpt.com/}} conversations conducted in Chinese. We consider these 500 ShareGPT samples as out-of-distribution (OOD) because the instructions originate from real-world applications without any modification or enhancement, and the implicit formats in responses significantly differ from those in our three data sources mentioned in Section \ref{sec:data-collection}. Analyzing the performance on these samples will be pivotal in assessing whether the skills learned with our OIFC-SFT dataset can generalize to both novel instructions and unseen response formats.

\paratitle{Evaluation Metrics.}
We evaluate the models' performances under the OIFC scenarios with two key metrics: helpfulness and format correctness. Both metrics are quantified on a scale from 0 to 1. The evaluation process involves automatically compiling the scoring rules, instructions, one-shot examples, ground-truth responses, and the model's generated responses into a structured evaluation prompt. This prompt is then used to invoke the external API to obtain evaluation results. The model employs greedy decoding with a repetition penalty set at 1.03 to generate responses. Details of the evaluation prompt are outlined in Table \ref{tab:eval_example}.

\begin{table}[h]
  \centering
  \caption{Example of the evaluation prompt we use in our OIFC benchmark.}
\footnotesize
\begin{CJK}{UTF8}{gbsn}{
\scalebox{0.83}{
    \begin{tabularx}{\textwidth}{c|X}
    \hline
    Chinese & \multicolumn{1}{m{10.5cm}}{请你需要为一个问题的回答打分。我会先给出一对示例问题和示例回答，然后提出待回答的问题。你需要对该问题的回答打分。分为两项，第一项是回答的正确性，即回答对解决问题是否有帮助，分数区间为0到1。第二项是回答是否遵循示例回答的格式规范，分数区间为0到1，0表示完全没有遵循，1表示完全遵循。以下是你需要打分的样本：\textbackslash n <json格式的问题、one-shot样例、标准答案、模型答案> \textbackslash n请先进行详细分析，最后将你的打分组织成以下格式：\textbackslash n\{'回答的正确性': 第一项分数, '回答的格式规范': 第二项分数\}}\\\hline
    English Translation & \multicolumn{1}{m{10.5cm}}{Please rate the answer to a question. I will first provide a pair of example questions and answers, then present the question to be rated. You need to score the answer to this question. There are two criteria: the first is the correctness of the answer, i.e., whether the answer is helpful in solving the problem, with a score range from 0 to 1. The second criterion is whether the answer follows the format standards of the example answers, with a score range from 0 to 1, where 0 indicates no adherence and 1 indicates full adherence. Below is the sample you need to rate:\textbackslash n
    <json format of the question, one-shot example, standard answer, model answer>\textbackslash n
    Please conduct a detailed analysis first, and then organize your scoring in the following format:\textbackslash n
    \{'Helpfulness': first criterion score, 'Format Correctness': second criterion score\}}\\
    \hline
    \end{tabularx}%

    }
  \label{tab:eval_example}%

}\end{CJK}
\end{table}%

\section{Experiments}
To evaluate the performance of large language models (LLMs) in One-shot Implicit Format Control (OIFC) settings, we conducted Supervised Fine-Tuning (SFT) using both 7B and 52B LLMs with the OIFC-SFT training data described previously. Initially, the pre-trained models are fine-tuned with conventional instruct data featuring open formats as \textit{stage-1}. This is followed by a \textit{stage-2} fine-tuning using OIFC-specific data.

\subsection{AF-7B}
AF-7B is our pre-trained foundation model with 7B parameters, developed following the techniques and experiences from the FLM \cite{teleflm,52b21t} series. This model has processed approximately 1.4T tokens and has achieved performance levels comparable to the Llama-2-7B \cite{llama-2}. In the stage-1 of fine-tuning, we use 50k math problems mixed with a minor proportion of open-domain questions, adhering to practices reported in FLM-2 \cite{teleflm}. This stage results in the AF-7B-Instruct, which serves as a baseline chat model trained without specific format constraints. For the second stage, we shift focus to fine-tuning with the OIFC-SFT dataset. The learning rate is initialized at 1e-5, and is decayed to 1e-9 over 4 epochs following a linear schedule. Batch size is set to 64; weight decay is set to 0.1. Optimization is performed using an AdamW \cite{adamw} optimizer with $\alpha_1=0.9$, $\alpha_2=0.95$, and a gradient clipping threshold of 1.0. The checkpoint after 1160 steps (2 epochs) is selected for evaluation, which is designated as AF-7B-OIFC.

The results on AF-7B are presented in Table \ref{tab:results} (top 2 rows). Initially, the AF-7B-Instruct model, fine-tuned during the first stage, exhibits poor performance in adhering to one-shot format examples. We find that the main issue is the model's tendency to ignore the specific format requirements, defaulting instead to the formats learned from the stage-1 SFT data. In contrast, after the second stage of fine-tuning with OIFC-SFT data, there is a notable improvement in performance—approximately 0.3 points on in-distribution (ID) formats and 0.15 points on out-of-distribution (OOD) formats. 

Interestingly, the baseline score for OOD formats is significantly higher than that for ID formats (0.71 vs. 0.62). This indicates that the implicit formats in our ID data is distinct from the widely-used GPT-like implicit formats, which is consistent with our intuition from case studies. Regarding the helpfulness scores, we observe slight improvement after stage-2 SFT. However, this is not noticeably perceptible to human annotators. 

Conclusively, our experiments demonstrate that OIFC-specific fine-tuning substantially enhances the model's ability to follow one-shot formats in AF-7B, while maintaining a level of helpfulness comparable to the initial training.

\subsection{FLM-2-52B}
To evaluate the efficacy on larger models, similar experiments were conducted on FLM-2-52B \cite{teleflm, 52b21t}. We began with FLM-2-52B-Instruct-2407 \footnote{\url{https://huggingface.co/CofeAI/FLM-2-52B-Instruct-2407}}, a chat model derived from the Tele-FLM-52B foundation and fine-tuned with carefully curated stage-1 SFT data. This model exhibits chat capabilities on par with leading proprietary models as assessed in the Alignbench evaluation \cite{alignbench}. In the second stage, FLM-2-52B-Instruct-2407 was fine-tuned using identical data and settings as those applied to the 7B model.

The results on FLM-2-52B are presented in Table \ref{tab:results} (bottom 2 rows). The increased capacity of the 52B model yields better results in format correctness, although these improvements remain unsatisfactory without subsequent OIFC fine-tuning. Our findings with the 52B model mirror those observed in the 7B experiments: OIFC fine-tuning crucially enhances format control capabilities without noticeably impacting the helpfulness of responses.

\begin{table}[htbp]
  \centering
  \caption{Experimental results.}
\footnotesize
\scalebox{1}{
    \begin{tabular}{l|cc|cc}
    \hline
    \multirow{2}{*}{Model} & \multicolumn{2}{c}{ID} & \multicolumn{2}{c}{OOD}\\\cline{2-5}
    & Helpfulness & Format & Helpfulness & Format\\\hline
    AF-7B-Instruct & 0.80 & 0.62 & 0.81 & 0.71 \\
    AF-7B-OIFC & 0.82 & 0.91 & 0.82 & 0.86 \\
    FLM-2-52B-Instruct & 0.91 & 0.71 & 0.92 & 0.81 \\
    FLM-2-52B-OIFC & 0.88 & 0.93 & 0.90 & 0.90 \\
    
    \hline
    \end{tabular}%

    }
  \label{tab:results}%
\end{table}%

\section{Related Work}
Controlled generation has been a persistent area of interest within the natural language processing community, particularly in summarization contexts \cite{faithfulness, global}. In the era of LLMs, controlling the model outputs is still a central focus for both the academic and the industrial communities \cite{we,zhang2023survey}. We roughly categorize prevailing methodologies into four main groups: (i) Controlling the decoding process with external constraints, e.g. automata \cite{efficient}, grammar \cite{grammar}, and guided search \cite{deal}. While these methods typically ensure high correctness, they often result in a decrease in the helpfulness of the responses. (ii) Supervised fine-tuning with data involving format constraints \cite{controlled-sft}. These methodologies are data-driven and suffers less performance degradation. However, the imposed format constraints are still hand-crafted, limiting their domain coverage. (iii) Constrained refining with the format-free responses from proprietary LLMs \cite{sketch}, relying on black-box models which may not always be transparent or adjustable. (iv) Few-shot in-context learning \cite{gpt3, cot, ditto}. In-context learning are feasible if models with strong capacities are available. However, constructing few-shot fine-tuning data for open-domain scenarios is challenging, and using more than one-shot examples significantly increases token usage, which can be inefficient in practical applications. We hope that our OIFC task formation and data to be complement to these existing methods.

\section{Conclusions}
In this work, we introduce the Open-domain Implicit Format Control (OIFC) framework to tackle the challenges of explicitly describing format requirements in real-world applications. We have developed a data collection pipeline that not only ensures diversity in the domains of the instructs and formats but also transforms low-quality responses into useful data. The resulting training dataset and testing benchmarks will be publicly available. Our experiments demonstrate that even basic fine-tuning settings on OIFC-SFT data can significantly enhance format control capabilities without noticeably affecting the original response quality. Looking ahead, future work will explore varied Supervised Fine-Tuning (SFT) configurations, such as integrating OIFC-SFT with other datasets in mixed SFT scenarios.

\bibliographystyle{plain}
\bibliography{custom1}

\end{document}